\documentclass[10pt,twocolumn,letterpaper]{article}

 \usepackage{cvpr}              
\usepackage{graphicx}
\usepackage{amsmath}
\usepackage{amssymb}
\usepackage{booktabs}
\usepackage{hyperref}
\usepackage{bm}
\usepackage{multirow}
\usepackage{makecell}

\hypersetup{colorlinks=true, urlcolor=red}
    
\def\paperID{13873} 
\def\confName{CVPR}
\def\confYear{2024}

\title{MureObjectStitch: Multi-reference Image Composition}

\author{Jiaxuan Chen$^1$, Bo Zhang$^1$, Qingdong He$^2$, Jinlong Peng$^2$, Li Niu$^1$\thanks{Corresponding author.} \\
 $^1$ MoE Key Lab of Artificial Intelligence, Shanghai Jiao Tong University\\
 $^2$ Youtu Lab, Tencent\\
{\tt \{chenjiaxuan,bo-zhang,ustcnewly\}@sjtu.edu.cn}, \\
{\tt \{yingcaihe,jeromepeng\}@tencent.com}
}

\begin{document}
\maketitle
\definecolor{cvprblue}{rgb}{0.21,0.49,0.74}

\begin{abstract}
Generative image composition aims to regenerate the given foreground object in the background image to produce a realistic composite image. The existing methods are struggling to preserve the foreground details and adjust the foreground pose/viewpoint at the same time. 
In this work, we propose an effective finetuning strategy for  generative image composition model, in which we finetune a pretrained model using one or more images containing the same foreground object. Moreover, we propose a multi-reference strategy, which allows the model to take in multiple reference images of the foreground object.   
The experiments on MureCOM dataset verify the effectiveness of our method. The code and model have been released \href{https://github.com/bcmi/MureObjectStitch-Image-Composition}{here} . 
\end{abstract}

\section{Introduction} \label{sec:intro}

The goal of generative image composition (object insertion) is inserting the foreground object into the background to produce a high-quality composite image. Typically, the model takes in a background image with specified foreground placement and a reference image of foreground object. In the generated image, the foreground is seamlessly blended into the background, with compatible illumination and geometry.  

The existing generative image composition methods can be roughly divided into high-authenticity methods and high-fidelity methods. As an example high-authenticity method, ObjectStitch~\cite{objectstitch} can generate realistic composite images, but the foreground details cannot be well preserved for those uncommonly seen objects or the objects with complex details. As an example high-fidelity method, ControlCom~\cite{zhang2023controlcom} designed a local enhancement module~\cite{zhang2023controlcom} to enrich the detail information. Such modification can greatly promote the foreground fidelity, but the model may inappropriately maintain the pose/viewpoint of foreground object, or generate images with distorted content structure and notable artifacts when attempting to adjust the pose/viewpoint. Therefore, based on the performance of previous methods \cite{objectstitch, zhang2023controlcom}, high authenticity and high fidelity can hardly be achieved at the same time. 

In this work, we propose a finetuning strategy when one or more images containing the same foreground object are available. Following \cite{objectstitch, zhang2023controlcom}, for each image, we segment the foreground object and perform illumination/geometry transformation to make it a reference image. The image after removing the foreground object becomes a background image. Based on pairs of background image and reference image, we can finetune the generative image composition model, similar to the training strategy used in previous works \cite{objectstitch, zhang2023controlcom}. 

\begin{figure*}[t]
\centering
\includegraphics[width=0.9\linewidth]{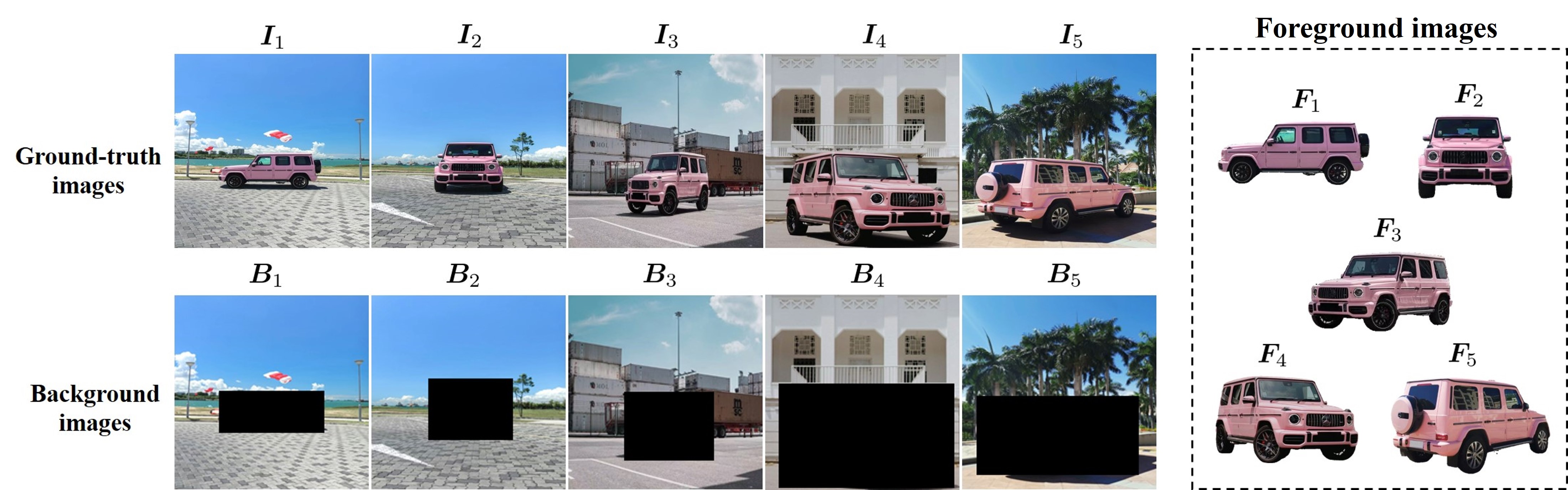} 
\caption {Illustration of ground-truth images, background images, and foreground images.}
\label{fig:multi_fg_imgs}
\end{figure*}

\begin{figure*}[t]
\centering
\includegraphics[width=0.85\linewidth]{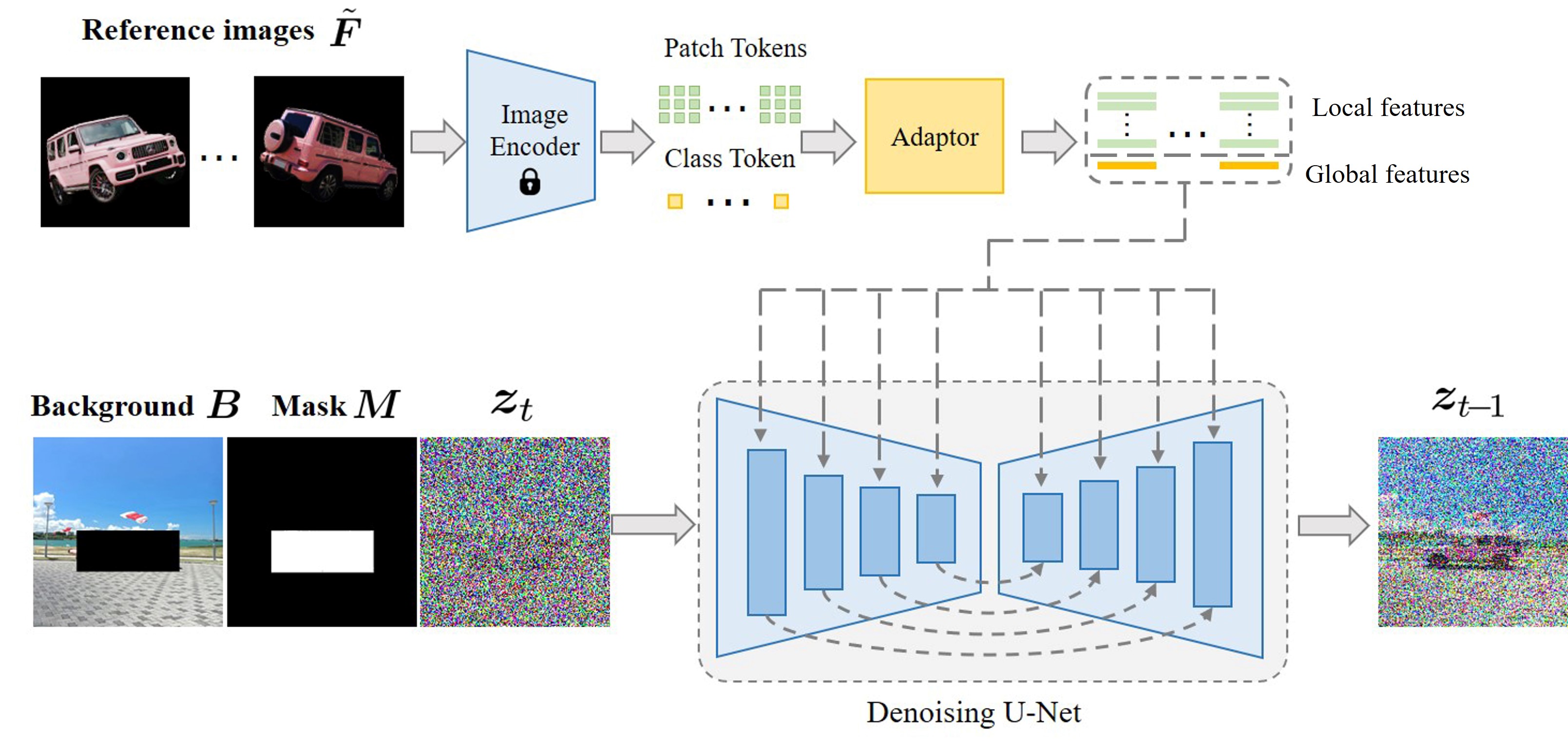} 
\caption {Illustration of our MureObjectStitch model.}
\label{fig:network}
\end{figure*}

Furthermore, we propose a multi-reference strategy, which allows the model to take in multiple reference images of foreground object. This is natural when we have more than one images containing the target object. In this case, we can take one image as the ground-truth image and segment the foreground objects in all images followed by illumination/geometry transformation as reference images.  
Using multiple reference images can enhance the model generation ability for diverse poses and viewpoints. Some existing generative composition methods \cite{objectstitch, zhang2023controlcom} extract the features from reference image and inject the reference features into denoising U-Net using cross-attention. This type of methods can be easily extended to multi-reference case by simply concatenating the reference features of multiple reference images. 

The combination of finetuning and multi-reference leads to our multi-reference finetuning strategy. We apply multi-reference finetuning to a high-authenticity method ObjectStitch \cite{objectstitch} and a high-fidelity method ControlCom \cite{zhang2023controlcom}, with experiments conducted on MureCom~\cite{lu2023dreamcom} dataset. 
We observe that for the high-authenticity method, the ability of detail preservation is dramatically improved, because the finetuned model can fit the details of target object. However, for the high-fidelity method, the ability of pose/viewpoint adjustment is still very weak. One possible reason is that the generation ability of the base model cannot be improved by finetuning on the target object. 
Therefore, we opt for applying multi-reference finetuning to high-authenticity method.  In particular, we choose ObjectStitch as base model and refer to its extension with \textbf{mu}lti-\textbf{re}ference finetuning strategy as MureObjectStitch.

\begin{figure*}[t]
\centering
\includegraphics[width=0.99\linewidth]{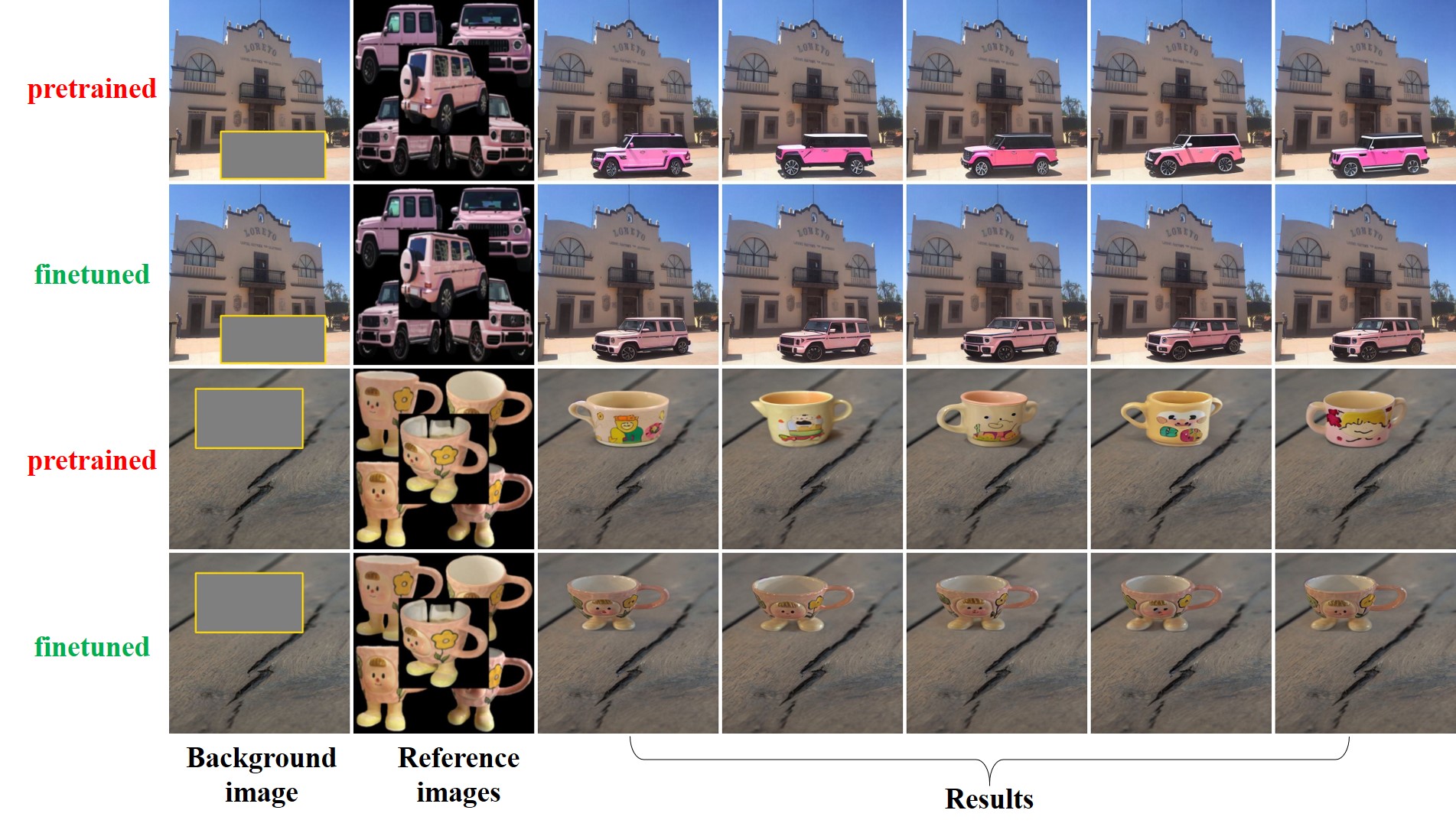} 
\caption {Visual comparison between pretrained ObjectStitch and our finetuned MureObjectStitch. In each example, from left to right, we show the background image with specified foreground placement, 5 reference images of foreground object, and 5 results using different random seeds. The results in odd rows are obtained using the pretrained ObjectStitch, and the results in even rows are obtained using the finetuned MureObjectStitch. }
\label{fig:results}
\end{figure*}

\section{Method} \label{sec:method}

We require a pretrained base model (\emph{e.g.}, ObjectStitch \cite{objectstitch}) and finetune it on a few images with the same target object. In Section~\ref{sec:pretraining}, we will briefly introduce the pretraining process of base model. In Section~\ref{sec:finetuning}, we will describe how to finetune the base model for a target object.

\subsection{Pretraining Base Model} \label{sec:pretraining}

Built upon the text-to-image Stable Diffusion (SD) model~\cite{rombach2022high}, ObjectStitch \cite{objectstitch} is retrained on a large-scale dataset Open Images~\cite{Kuznetsova2020TheOI}, which has annotated bounding boxes for foreground objects. For each image $\bm{I}$, we remove the content within foreground bounding box and get the background image $\bm{B}$. Then, we segment the foreground object followed by illumination and geometry perturbation, yielding the reference image $\tilde{\bm{F}}$. 
 
ObjectStitch modifies the input of denoising U-Net as the concatenation of background $\bm{B}$, bounding box mask $\bm{M}$, and noisy latent $\bm{z}_t$, as illustrated in Figure~\ref{fig:network}. The reference image $\tilde{\bm{F}}$ is passed through a pretrained CLIP image encoder \cite{CLIPscore} and an adaptor. The extracted global feature and local features of this reference image are injected into the denoising U-Net via cross-attention. Initialized with the pretrained SD model weights, the whole denoising U-Net is retrained to predict the noise.

After retraining on Open Images dataset~\cite{Kuznetsova2020TheOI}, the obtained model can adjust the pose/viewpoint to match the background and generate the high-quality composite images. However, the foreground details cannot be well preserved.

\subsection{Finetuning for Target Object} \label{sec:finetuning}
Assuming that we have $N$ images $\{\bm{I}_i|_{i=1}^N\}$ containing the target object, we extract the bounding boxes of target objects and obtain the bounding box masks $\{\bm{M}_i|_{i=1}^N\}$. After erasing the content within bounding boxes, we obtain the background images $\{\bm{B}_i|_{i=1}^N\}$ with specified foreground placement. Based on the extracted foreground bounding boxes, we segment the foreground images $\{\bm{F}_i|_{i=1}^N\}$. Each foreground image is tweaked using perspective transformation and color transfer, leading to the reference images $\{\tilde{\bm{F}}_i|_{i=1}^N\}$. The original images $\{\bm{I}_i|_{i=1}^N\}$ serve as the ground-truth images. The ground-truth images, background images, and foreground images are shown in Figure~\ref{fig:multi_fg_imgs}. 

For each background image, we use $N$ reference images and expect the generated image to approach the ground-truth image.  
The only difference between MureObjectStitch and ObjectStitch~\cite{objectstitch} is that we concatenate the features (both global features and local features) of all reference images, which are injected into the denoising U-Net via cross-attention.  
Given a variety of reference features, the model can learn to select the most relevant features conditioned on the contextual information in the background image.
We finetune the base model in Section~\ref{sec:pretraining} using $N$ pairs of background images and ground-truth images.

After finetuning the model, given a new background image with specified foreground placement, we start from random Gaussian noise $\bm{z}_T\sim \mathcal{N}(0,1)$ and produce the composite image through the denoising process. 

\begin{figure*}[t]
\centering
\includegraphics[width=0.89\linewidth]{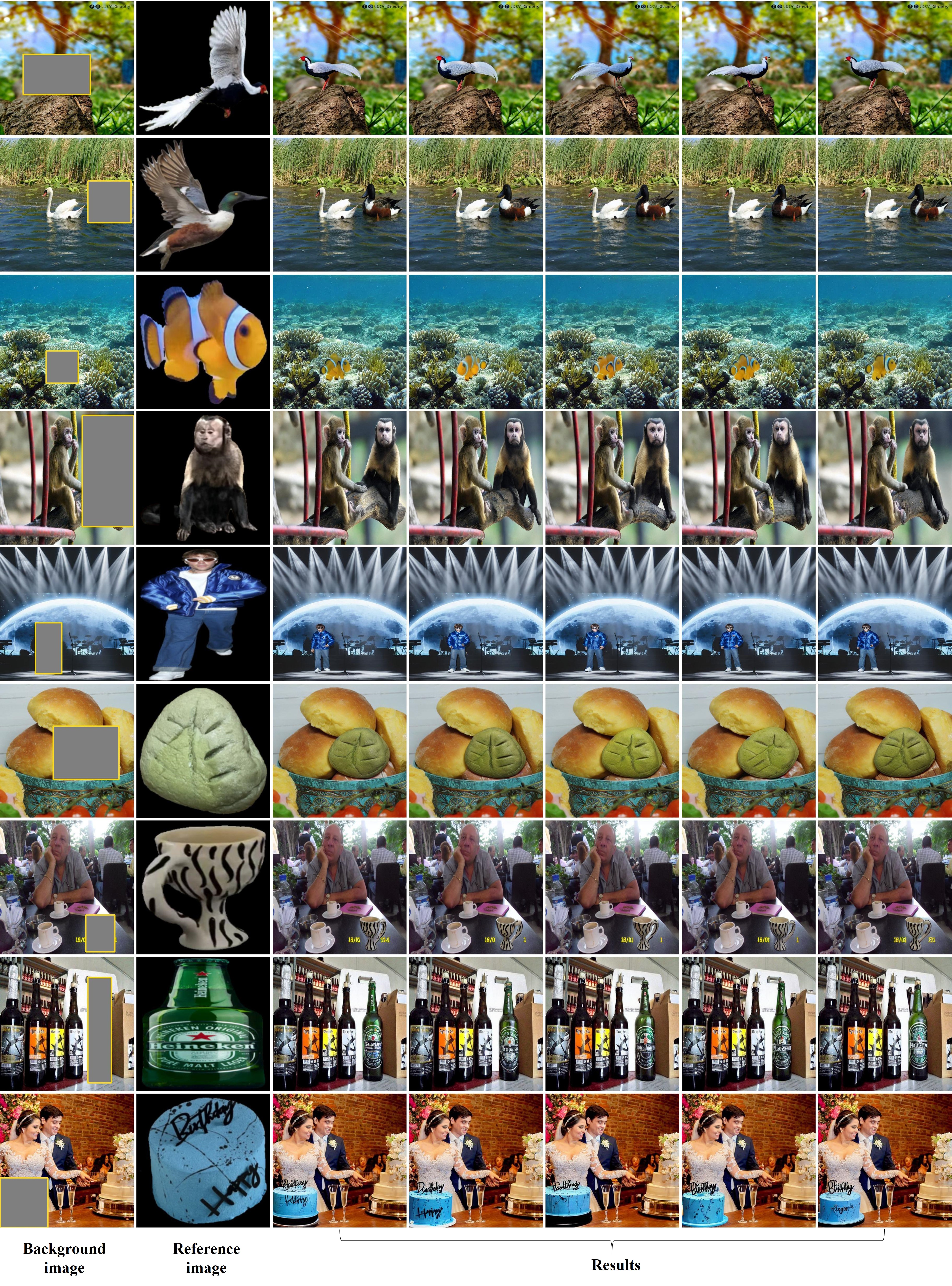} 
\caption {Visual results of our finetuned MureObjectStitch. In each example, from left to right, we show the background image with specified foreground placement, one example reference image of foreground object, and 5 results using different random seeds. }
\label{fig:more_results1}
\end{figure*}

\begin{figure*}[t]
\centering
\includegraphics[width=0.89\linewidth]{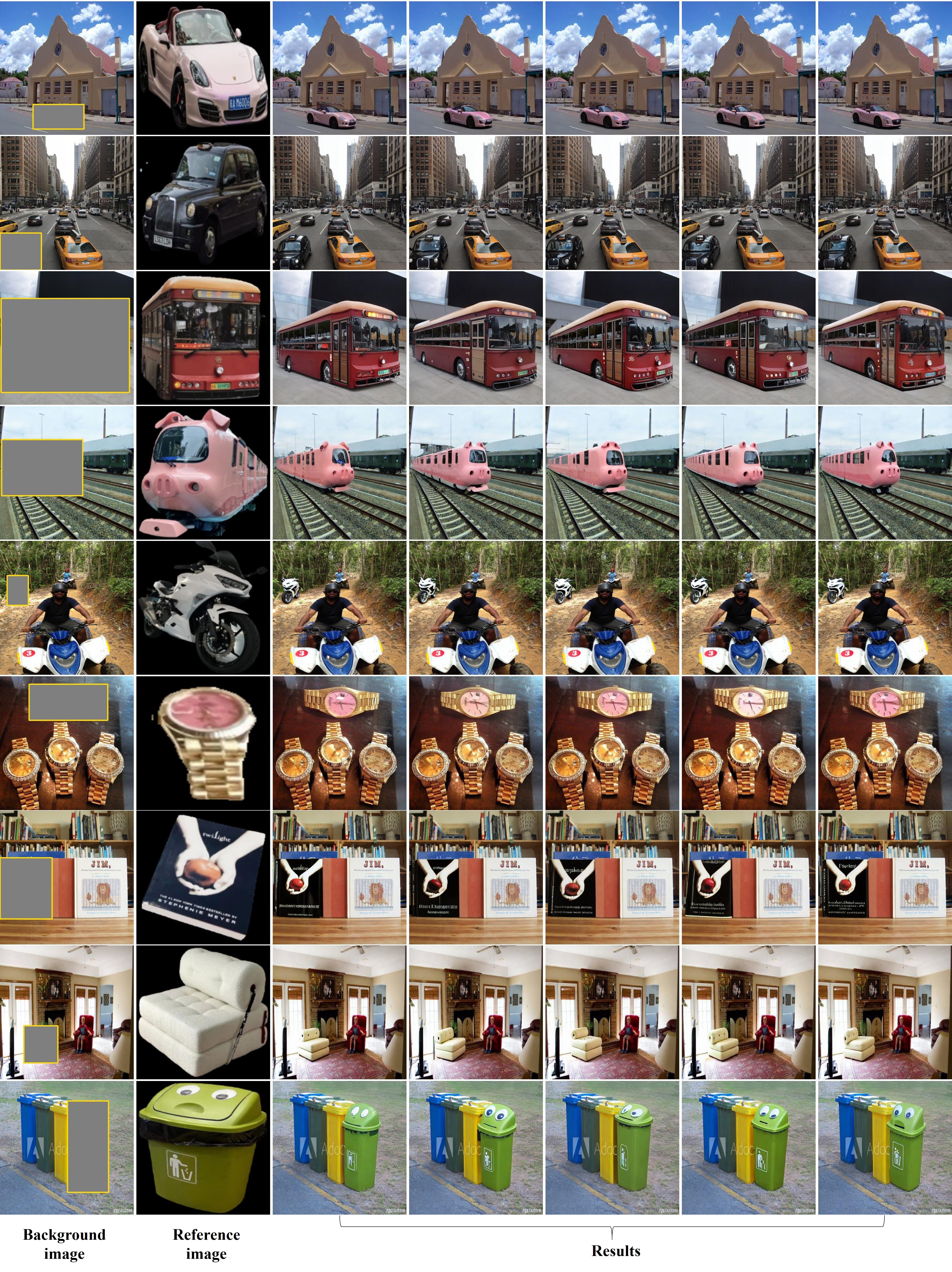} 
\caption {Visual results of our finetuned MureObjectStitch. In each example, from left to right, we show the background image with specified foreground placement, one example reference image of foreground object, and 5 results using different random seeds. }
\label{fig:more_results2}
\end{figure*}

\section{Experiment} \label{sec:exp}

We conduct experiments on MureCom \cite{lu2023dreamcom} dataset.
In MureCom \cite{lu2023dreamcom} dataset, there are $32$ foreground categories and each foreground category is associated with $20$ background images.
Each background image has a manually annotated bounding box suitable for placing the object from the corresponding foreground category.
Each foreground category has $3$ unique foreground objects with $5$ images for each foreground object. 

For each foreground object, we finetune the model based on $5$ images. During testing, we evaluate the model on 20 background images corresponding to its foreground category. For each foreground object, we finetune the model for 150 epochs, which can generally achieve satisfactory results. In some cases, finetuning more epochs (\emph{e.g.}, 200 epochs) is helpful for keeping more details, yet at the risk of distorted content and improper illumination. Finetuning 150 epochs takes about 15 minutes on a single A6000 GPU. 

The visual results are shown in Figure~\ref{fig:results}. In each example, from left to right, we show the background image with specified foreground placement, 5 reference images of foreground object, and 5 results using different random seeds. The results in odd rows are obtained using the pretrained ObjectStitch, and the results in even rows are obtained using the finetuned MureObjectStitch. It is evident that our finetuned MureObjectStitch can better preserve the details of foreground object while maintaining the ability to adjust the illumination and geometry of foreground object according to the background. 

We also show more results of MureObjectStitch in Figure~\ref{fig:more_results1} and \ref{fig:more_results2}.  In each example, from left to right, we show the background image with specified foreground placement, one example reference image of foreground object, and 5 results using different random seeds. It can be seen that MureObjectStitch can generally achieve compelling results for a wide range of foreground categories (\emph{e.g.}, animal, person, vehicle, food, furniture). In the generated images, the foreground object interacts naturally with the background and other objects (\emph{e.g.}, bird and monkey in Figure~\ref{fig:more_results1}, train and waste container in Figure~\ref{fig:more_results2}). The illumination and geometry of generated foreground object are compatible with the background. Our method can strike a good balance between high authenticity and high fidelity, yielding the foreground object with well-adjusted pose/viewpoint and sufficient faithful details.

\section{Conclusion} \label{sec:conclusion}
In this work, we have proposed a simple yet effective multi-reference finetuning strategy for  generative image composition model. Our model supports an arbitrary number of reference images, which enables great flexibility in practical usage. 
The experimental results on MureCOM dataset have demonstrated that our model can preserve the foreground details and adjust the foreground pose/viewpoint at the same time, which has never been achieved before. 

{
    \small
    \bibliographystyle{ieeenat_fullname}
    \bibliography{main.bbl}
}


\end{document}